\documentclass[letterpaper]{article} % DO NOT CHANGE THIS
\usepackage{aaai25}  % DO NOT CHANGE THIS
\usepackage{times}  % DO NOT CHANGE THIS
\usepackage{helvet}  % DO NOT CHANGE THIS
\usepackage{courier}  % DO NOT CHANGE THIS
\usepackage[hyphens]{url}  % DO NOT CHANGE THIS
\usepackage{graphicx} % DO NOT CHANGE THIS
\usepackage{natbib}  % DO NOT CHANGE THIS AND DO NOT ADD ANY OPTIONS TO IT
\usepackage{caption} % DO NOT CHANGE THIS AND DO NOT ADD ANY OPTIONS TO IT
\usepackage{algorithm}

\usepackage[utf8]{inputenc}
\usepackage[T1]{fontenc}
\usepackage{latexsym}
\usepackage{amssymb}
\usepackage{amsmath}
\usepackage{amsthm}
\usepackage{multirow}
\usepackage{booktabs}
\usepackage{enumitem}
\usepackage{graphicx}
\usepackage[dvipsnames,svgnames,x11names,table]{xcolor}
\newcommand{\dorowcolors}{\rowcolors{2}{gray!15}{white}}
\usepackage{caption}
\usepackage{algorithm}
\usepackage{algpseudocode}

\usepackage{microtype}
\usepackage{newfloat}
\usepackage{listings}
\DeclareCaptionStyle{ruled}{labelfont=normalfont,labelsep=colon,strut=off} % DO NOT CHANGE THIS
\floatstyle{ruled}
\newfloat{listing}{tb}{lst}{}
\floatname{listing}{Listing}
\title{A Knowledge Distillation-Based Approach to Enhance Transparency of Classifier Models}
\author{
    Yuchen Jiang\equalcontrib,
   Xinyuan Zhao\equalcontrib,
   Yihang Wu,
    Ahmad Chaddad\equalcontrib
}
\affiliations{Artificial Intelligence for Personalised Medicine, School of Artificial Intelligence, Guilin University of Electronic Technology\\    Jinji Road, Guilin 541004, China\\
    ahmad8chaddad@gmail.com
}

\usepackage{bibentry}
\begin{document}

\maketitle

\begin{abstract}
With the rapid development of artificial intelligence (AI), especially in the medical field, the need for its explainability has grown. In medical image analysis, a high degree of transparency and model interpretability can help clinicians better understand and trust the decision-making process of AI models. In this study, we propose a Knowledge Distillation (KD)-based approach that aims to enhance the transparency of the AI model in medical image analysis. The initial step is to use traditional CNN to obtain a teacher model and then use KD to simplify the CNN architecture, retain most of the features of the data set, and reduce the number of network layers. It also uses the feature map of the student model to perform hierarchical analysis to identify key features and decision-making processes. This leads to intuitive visual explanations. We selected three public medical data sets (brain tumor, eye disease, and Alzheimer's disease) to test our method. It shows that even when the number of layers is reduced, our model provides a remarkable result in the test set and reduces the time required for the interpretability analysis.
%here i correct comment 1(remove the link of code)

\end{abstract}
\begin{links}
\link{Code}{https://github.com/AIPMLab/KD-FMV}
\end{links}

\section{Introduction}
Artificial intelligence (AI) is now widely used in various fields, especially deep learning (DL), which advances DL applications in more autonomous directions \cite{wang2023scientific, al2024deep}. In medical imaging, AI has become integral to healthcare \cite{ chaddad2023federated, amiri2024deep}. Despite Convolutional Neural Networks (CNNs) performing well in medical classification tasks, they lack transparency, particularly in decision-making \cite{chaddad2024generalizable,hassija2024interpreting}. This is crucial in medicine because unexplainable AI results cannot convince physicians and may lead to medical accidents \cite{jin2023guidelines}. Explainable artificial intelligence (XAI) addresses this by enhancing model transparency and understandability, helping physicians trust AI results and improving medical decisions \cite{chaddad2023explainable, van2022explainable}.

DL models often have large parameters, complicating interpretation. Knowledge distillation (KD) transfers knowledge from complex to simpler models \cite{hinton2015distilling}, widely used in medical tasks. For instance, \cite{wang2023ssd} proposed a self-supervised method integrating diverse knowledge into KD for skin disease classification, achieving higher performance. The Distilled Student Network, using KD, detects melanoma from dermoscopic images more efficiently than pre-trained models \cite{khan2022knowledge}. Smaller models are computationally cheaper and easier to understand. Our key question is: \textit{How to design a simple model with feasible classification accuracy while maintaining high interpretability?} This study simplifies CNN architecture through KD, retaining most dataset features and reducing network layers. Our approach involves: 1) training a teacher model such as CNNs and performing KD to obtain a simpler student model, 2) using average feature maps on limited layers for intuitive visualization. We use Kullback-Leibler (KL) divergence as a loss function to preserve useful features while simplifying the model. Feature distillation aims to make the student model learn similar feature representations as the teacher model, enhancing student model performance. We performed a layer-by-layer interpretability analysis, reducing CNN layers without compromising effectiveness. Analyzing each student model layer's feature map identifies key features and decision-making processes. Our goal is to maximize model performance and interpretability while maintaining a lightweight structure. To verify the proposed approach, we used two interpretability methods: 1) Grad-CAM, highlighting image regions contributing to model predictions by visualizing the last convolutional layer \cite{selvaraju2017grad}, and 2) SHapley Additive exPlanations (SHAP), assigning each feature a contribution value to the model output \cite{lundberg2017unified}. Our contributions are:
\begin{itemize}
\item We propose a solution to the interpretability problem using KD in the context of DL, ensuring high performance and quick explanations.
\item We simulate the proposed model using three medical datasets to demonstrate classifier accuracy with few layers.
\item We compare the proposed interpretability model with the Grad-CAM and SHAP models using both visual explanations (e.g., heat map) and quantitative metrics (e.g., Fidelity score). 
\end{itemize}

\section*{Related Work}\label{sec:2}
\textit{Knowledge Distillation}: Several KD methods are used today in medical applications \cite{gou2021knowledge}. For example, a study proposes a KD strategy for building compact DL models suitable for the classification of chest X-ray images (CXR) with multiple labels in real time. It uses different CNNs and transformers as teacher networks to transfer KD to smaller student networks \cite{termritthikun2023explainable}. In \cite{park2022self}, distillation for self-supervision and self-training demonstrates higher performance in diagnosing tuberculosis, pneumothorax, and COVID-19 even when using unlabeled data. In \cite{patel2023logistic}, the study predicted results using the Open Access Serial Imaging Study (OASIS) MRI dataset. 
The process includes data exploration, data preprocessing, and the hybrid model that integrates both logistic regression and decision tree algorithms. The proposed hybrid model outperforms existing models, with an overall accuracy of 96\%. Furthermore, in \cite{liu2023segmentation}, 
they introduced a dual-branch architecture that improves performance by transferring knowledge from a teacher to a student model, achieved by minimizing Shannon entropy and KL divergence. This method achieved outstanding results in a public data set for the left ventricular endocardium segmentation task.

\noindent\textit{Explainable AI}: As previously presented, the importance of XAI in medical diagnosis continues play a major role in providing clear interpretability \cite{chaddad2023survey}. For example, in \cite{raihan2023detection}, the authors used Extreme Gradient Boosting (XGboost) to predict whether a patient has chronic kidney disease (CKD). SHAP analysis is used to explain the impact of features on the XGBoost model. For example, using SHAP analysis and the Biogeography-Based Optimization (BBO) algorithm, hemoglobin models arebumin contributed mainly to the detection of CKD. In \cite{alabi2023machine}, machine learning (ML) models are used with understandable AI to build a prognostic model to group patients with nasopharyngeal cancer (NPC) into two groups based on their survival probabilities, high and low. Local Interpretable Model Agnostic Explanations (LIME) and SHAP models were used to provide interpretability. LIME and SHAP models identify personalized protective and risk factors for each NPC patient, revealing new non-linear relationships between input features and survival odds. Similarly, Grad-CAM introduced a novel XAI framework that enhances feature explainability for decision making in tumor diagnosis using ultrasound data, as proposed in \cite{song2023new}. This framework is capable of identifying regions that are more relevant and feature-associated compared to the traditional Grad-CAM. Currently, an automated approach for detecting and classifying leukemia was demonstrated in a data set independent of the subject, using deep transfer learning supported by Grad-CAM visualization \cite{abhishek2023automated}. Unlike previous methods, our approach uses KD to reduce the number of CNN layers and outputs the main features of each layer to improve the interpretability analysis. Furthermore, the model can fully demonstrate the decision-making process while retaining most of the key features. 

\section*{Method}\label{sec:3}

\begin{figure*}[htp]
     \centering
    \includegraphics[width=1\linewidth]{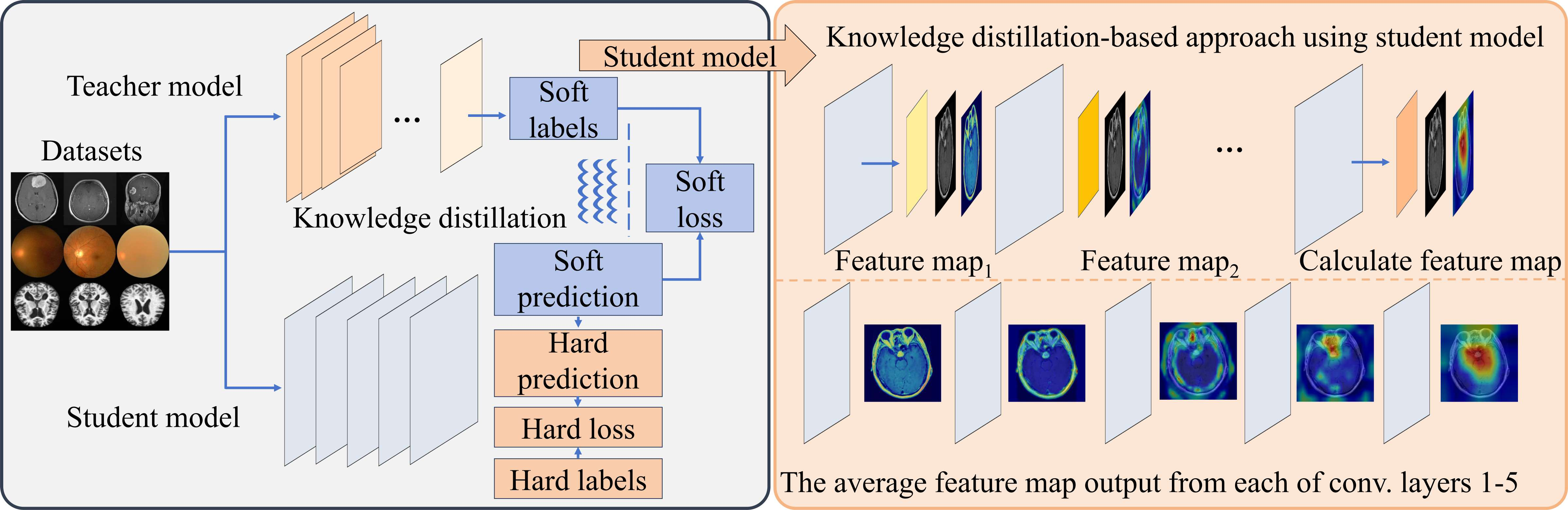}
    \caption{Flowchart of distilling knowledge and explainable AI.  1) distilling knowledge from DenseNet121 to a smaller model; then, conducting interpretability analysis. It starts by training DenseNet121, and then it distills knowledge into a five-layer custom CNN by minimizing the loss. 2) It obtains feature maps layer by layer and computes their averages, and 3) through color mapping, it identifies the key parts of the image that influence the model decision.}
    \label{pipeline}

\end{figure*}

Figure \ref{pipeline} illustrates the pipeline of our approach.
The KD method is used to train CNN, preserving its essential features, and the average feature map is used to illustrate the decision-making process at each layer, emphasizing the decision-making process within CNN. Algorithm \ref{ALG:1} provides a basic procedure of our approach.

\noindent\textbf{KD.} The objective of KD is to have the student model learn to mimic the behavior of the teacher model. This is typically achieved through a combination of the following two types of loss functions. The first one is \textbf{Hard Loss} (denoted as $\ell_{HL}$), the $\ell_{HL}$ function we used is the Cross-Entropy Loss, which for a multi-class classification problem. $M$ represents the total number of classes, \( N \) represents the total number of samples, and \( y_{S,i,c} \) denotes the predicted probability of the student model for the \( i \)-th sample belonging to class \( c \). $y_{i, c}$ represents that for the $i$-th sample, the true label for class $c$. In one-hot encoding, if sample $i$ belongs to class $c$, then $y_{i, c}=1$; otherwise, it is 0. $\log \left(y_{S, i, c}\right)$ is the logarithm of the predicted probability. The log function amplifies the penalty for probabilities below 1, making the model focus more on those samples that are predicted incorrectly. It can be expressed as follows.
\begin{equation}\label{3.1}\footnotesize
\ell_{H L}=-\frac{1}{N} \sum_{i=1}^N \sum_{c=1}^M y_{i, c} \log \left(y_{S, i, c}\right)
\end{equation}
\begin{algorithm}[!h]\footnotesize
\caption{Knowledge Distillation and Feature Map Visualization (KD-FMV).}
\begin{algorithmic}

\Require

\Statex $\mathcal{D}_t$: Target data set

\Statex $E$: Total epochs for student training

\Statex $P_{\text{teacher}}$: Path to pre-trained teacher model
\Statex $T$: Temperature for distillation
\Statex $\alpha$: Weight to balance hard and soft loss

\Ensure Visualize feature maps for an image
\State 
% \Statex Optimally trained student model and visualized feature maps

\State Initialize  $\mathcal{M}_T \leftarrow P_{\text{teacher}}$, $\mathcal{M}_S$
\label{code}
    % \State \Return $\mathcal{L}_{distill}$
\State 
% \State Preprocess $\mathcal{D}_t$ into datasets
\For{$epoch = 1$ to $E$}
    \For{each batch $(x, y)$}
        \State $y_{T} \gets \mathcal{M}_T(x)$ \Comment{\textcolor{ForestGreen}{Obtain predicts}}
        \State Update $\mathcal{L}_{distill} \gets \alpha \ell_{HL} + (1-\alpha) \ell_{SL}$ \Comment{\textcolor{ForestGreen}{Eq. \ref{3.3}}}
        \State Update $\mathcal{M}_S$ \Comment{\textcolor{ForestGreen}{Eq. \ref{3.3}}}
    \EndFor
    \State Evaluate and potentially save $\mathcal{M}_S$
\EndFor
\State 

\For{each layer in $\mathcal{M}_S$}
        \State $F \gets \text{Extract feature maps for } image$
        \State Obtain $A(i, j)=\frac{1}{N} \sum_{k=1}^N F_k(i, j)$ \Comment{\textcolor{ForestGreen}{Eq. \ref{3.6}}}
    \EndFor
    
% \State 
% \State Load and visualize feature maps for an image

\end{algorithmic}
\label{ALG:1}
\end{algorithm}
\noindent\textbf{Soft Loss} ($\ell_{SL}$) is the loss between the softened output of the teacher model and the softened version of the true labels. It differs from traditional KD by directly comparing the output of the teacher model with the adjusted real labels. This guides the student model to achieve the performance of the teacher model more directly.
The temperature parameter $T$ adjusts the "softness" of the probabilities. A higher temperature results in a softer probability distribution, reducing the differences between the largest and smallest values. This softening highlights the relative confidence of predictions rather than absolute values, which is useful for transferring knowledge.
The softmax function is applied to the outputs \(\frac{y_{\text{true}, c}}{T}\) and \(\frac{y_{T, c}}{T}\), where: \(y_{\text{true}, c}\) is the true label for class \(c\), typically represented as a one-hot encoded vector. \(y_{T, c}\) is the logit output of the teacher model for class \(c\). The number of classes is denoted by \(M\). The soft loss \(\ell_{SL}\) is formulated as:
\begin{equation}\label{3.2}
\ell_{SL}\left(y_{\text{true}}, y_T, T\right) = -\sum_{c=1}^M \text{\fontsize{5}{6}\selectfont $\left(\operatorname{softmax}\left(\frac{y_{\text{true}, c}}{T}\right)\right) \log \left(\operatorname{softmax}\left(\frac{y_{T, c}}{T}\right)\right)$}
\end{equation}

where, the logarithm is applied to the softmax probabilities, which is a component of the cross-entropy loss. The loss \(\ell_{SL}\) effectively computes the cross-entropy between \(\operatorname{softmax}\left( \frac{y_{\text{true}, c}}{T} \right)\) and \(\operatorname{softmax}\left( \frac{y_{T, c}}{T} \right)\), which corresponds to minimizing the KL divergence when \(\operatorname{softmax}\left( \frac{y_{\text{true}, c}}{T} \right)\) is treated as the ground-truth distribution.

The total loss function is a weighted sum of $\ell_{HL}$ and $\ell_{SL}$. $\alpha$ is a weight factor that is used to balance $\ell_{HL}$ and $\ell_{SL}$. At each training step, the predictive output of the teacher model is first calculated. Then, the student model is run within a gradient tape, the combined loss function is calculated, and the student model weights are updated using backpropagation (i.e., gradient descent). It could be shown:

\begin{equation}\label{3.3}
 \mathcal{L}_{distill} =\alpha \cdot \ell_{HL} +(1-\alpha) \cdot \ell_{SL}
\end{equation}

Temperature-scaled soft loss adjusts the softmax output of both the teacher and student models to produce a softer probability distribution over classes. This is achieved by dividing the logits (the inputs to the softmax function) by a temperature parameter $T$. A higher temperature produces a softer probability distribution, meaning that the differences between the probabilities of each class are less pronounced. $z_i$ is the logit for class $i$, and $T$ is the temperature parameter, with $T>1$ to produce a softer distribution.
\begin{equation}\label{3.4}\footnotesize
\operatorname{Softmax}_T\left(z_i\right)=\frac{e^{z_i / T}}{\sum_j e^{z_j / T}}
\end{equation}

When we obtain the CNN model after KD, the activation map (feature map) of the convolutional layer of the trained CNN is visualized. These visualizations can help to understand the parts of an image that activate certain features in the neural network involved in the decision-making process of the model. The feature map is extracted for a given convolutional layer by forwarding the input image through the network to that layer. If the convolutional layer has multiple filters, each filter produces a separate 2D activation map for the same input image. These are stacked along the depth dimension. The activation of a filter $f$ in a convolutional layer at a position $(i, j)$ for a given input image can be mathematically represented as the convolution of the input image $I$ with filter weights $W_f$, followed by the addition of a bias term $b_f$, and the application of a nonlinear activation function $\sigma$:
\begin{equation}\label{3.5}
\operatorname{Activation}_f(i, j)=\sigma\left(\left(I * W_f\right)(i, j)+b_f\right)
\end{equation}

\noindent As the number of filters increases, especially in deep CNN, it becomes impractical to show each feature map individually, as this can lead to information overload and make it challenging to gain intuitive insights. To solve this, we adopt the method of averaging feature maps. Specifically, we averaged the pixel values of all feature maps in the same layer at the corresponding positions to generate a single average feature map. The mathematical expression for calculating the average feature map $A$ for a given layer with $N$ filters is as follows:
\begin{equation}\label{3.6}\footnotesize
A(i, j)=\frac{1}{N} \sum_{k=1}^N F_k(i, j)
\end{equation}

\noindent where $A(i, j)$ represents the value of the average feature map at position $(i, j), F_k(i, j)$ is the activation value at position $(i, j)$ in the feature map produced by the $k$-th filter, and $N$ is the total number of filters in the layer. The resulting average feature map can then be visualized directly or further processed, such as overlaying it on the original input image to see which regions contribute the most to the activations in that layer.

\section*{Experiment}\label{sec:4}
% We simulate our approach using three public datasets as follows. 

\subsection*{Datasets}
\noindent\textit{Brain tumor}: This dataset \cite{nickparvar2021brain} is a collection that combines three separate data sets: figshare \cite{BrainTumorDataset}, the SARTAJ \cite{BrainTumorClassification}, and Br35H \cite{br35h2020}. A total of 7023 MRI images of the human brain are included. These images are classified into four classes: glioma (n=1621), meningioma (n=1645), no tumor (n=2000), and pituitary (n=1757). Unlike other datasets, official website has predefined test and training sets for brain tumor dataset. Therefore, our training-to-validation ratio is 8:2.

\noindent\textit{Eyes disease}: This dataset consists of 4217 Colour fundus photography (CFP) with four classes: cataract (n = 1038), diabetic retinopathy (n = 1098), glaucoma (n = 1007) and normal images (n = 1074)  \cite{doddi2023eye}. We randomly divided the images according to training, validation and testing with a ratio of 7:2:1. 

\noindent\textit{Alzheimer}: The dataset consists of MRI images grouped into four classes: Mild (n = 8960), moderate (n = 6464), non-demented (n = 9600), and very mild (n=8960) \cite{AugmentedAlzheimerMRIDataset}. The dataset contains both augmented images (processed through random rotations, cropping, color transformations, etc.) and original images. The images are randomly divided into training, validation, and testing ratios of 7:2:1. For detailed information on the training/validation and testing sets, refer to Table \ref{number}.

\begin{table}[!ht]\scriptsize
    \centering
    \setlength{\tabcolsep}{6.7pt}
    \renewcommand{\arraystretch}{0.8}
    \dorowcolors
    \resizebox{\columnwidth}{!}{
    \begin{tabular}{ccccc}
        \toprule
         Name & Train & Validation & Test  & Total \\
           \midrule
     \rowcolor{white}\multicolumn{5}{c}{\textit{Brain tumor}} \\
     \midrule
     Glioma & 1057 & 264 & 300 & 1621\\
     Meningioma & 1071 & 268 & 306 & 1645 \\
     No tumor & 1276 & 319 & 405 & 2000\\
     Pituitary & 1167 & 290 & 300 & 1757\\
    \midrule
    \rowcolor{white}\multicolumn{5}{c}{\textit{Eye disease}} \\
    \midrule
    Cataract & 582 & 145 & 311 & 1038\\
    Diabetic retinopathy & 615 &154 & 329& 1098\\
    Glaucoma & 564 & 141 & 302 & 1007\\
    Normal & 602 & 150 & 322 & 1074\\
    \midrule
    \rowcolor{white}\multicolumn{5}{c}{\textit{Alzheimer}} \\
    \midrule
    Mild demented & 5734 & 1794 & 1792 & 9320 \\
    Moderate demented & 4136 & 1034 & 1294 & 6464\\
    Non demented &6144 & 1536 & 1920 & 9600 \\
    Very mild demented & 5734 & 1434 & 1792 & 8960\\
    \bottomrule
    \end{tabular}
    }
  \caption{Number of samples for training, validation, and test sets for the three datasets.}
    \label{number}\vspace{-0.3cm}
\end{table}

\begin{figure}[h]
     \centering
    \includegraphics[width=1.0\linewidth]{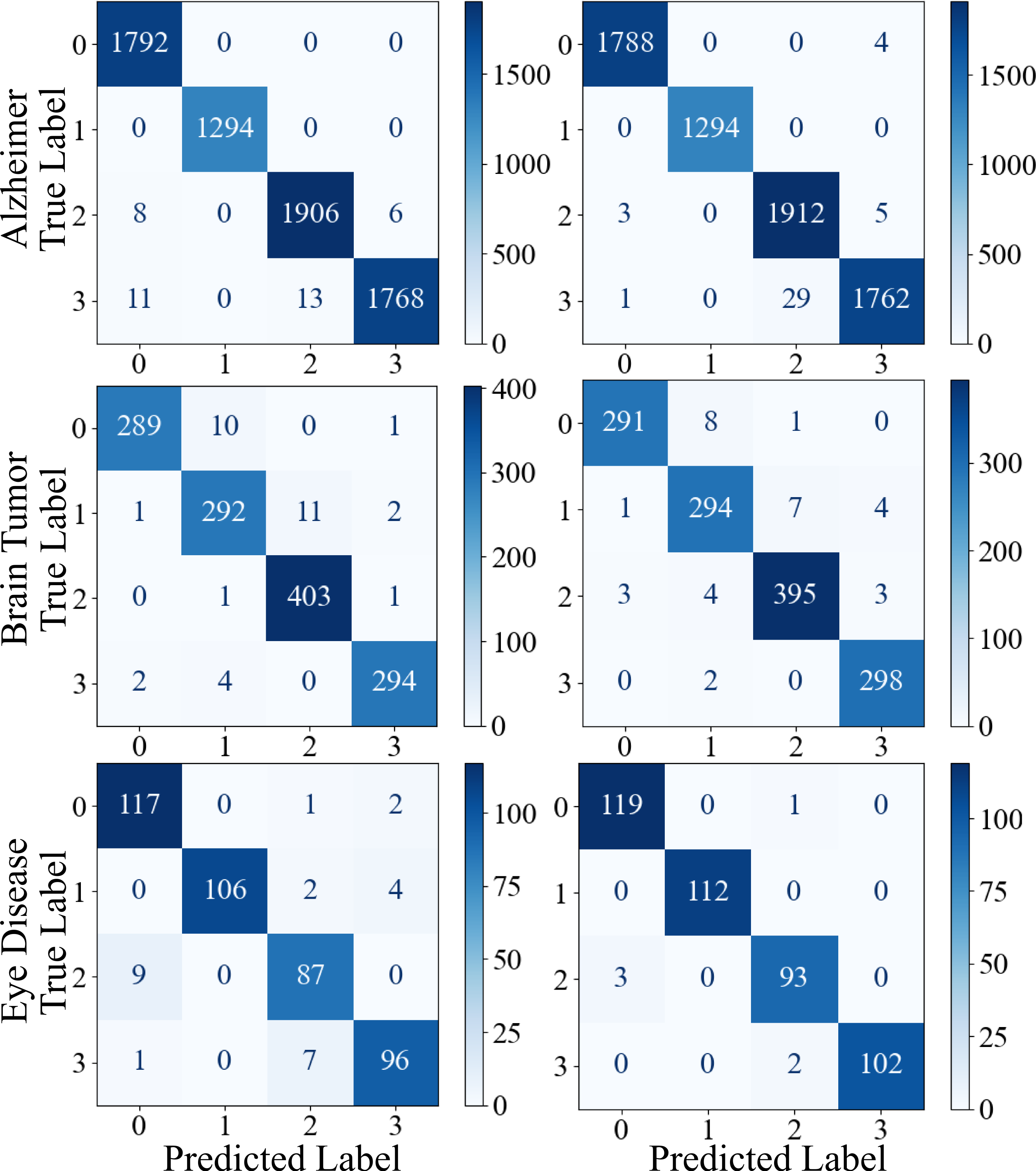}
    \caption{Confusion matrix of the student (\textbf{Left}) and teacher (\textbf{Right}) model on brain tumor, eyes disease and alzheimer datasets, respectively. In brain tumor dataset, 0, 1, 2, and 3 represent Glioma tumor, Meningioma, No tumor and Pituitary, respectively. In alzheimer dataset, 0, 1, 2 and 3 indicate Mild demented, Moderate demented, Non demented and Very mild demented, respectively. In Eye-disease data set, 0, 1, 2 and 3 denote Cataract, Diabetic retinopathy, Glaucoma and Normal, respectively.}
    \label{Fig:CM}
\end{figure}  

\begin{figure*}[!t]
     \centering
    \includegraphics[width=0.82\textwidth]{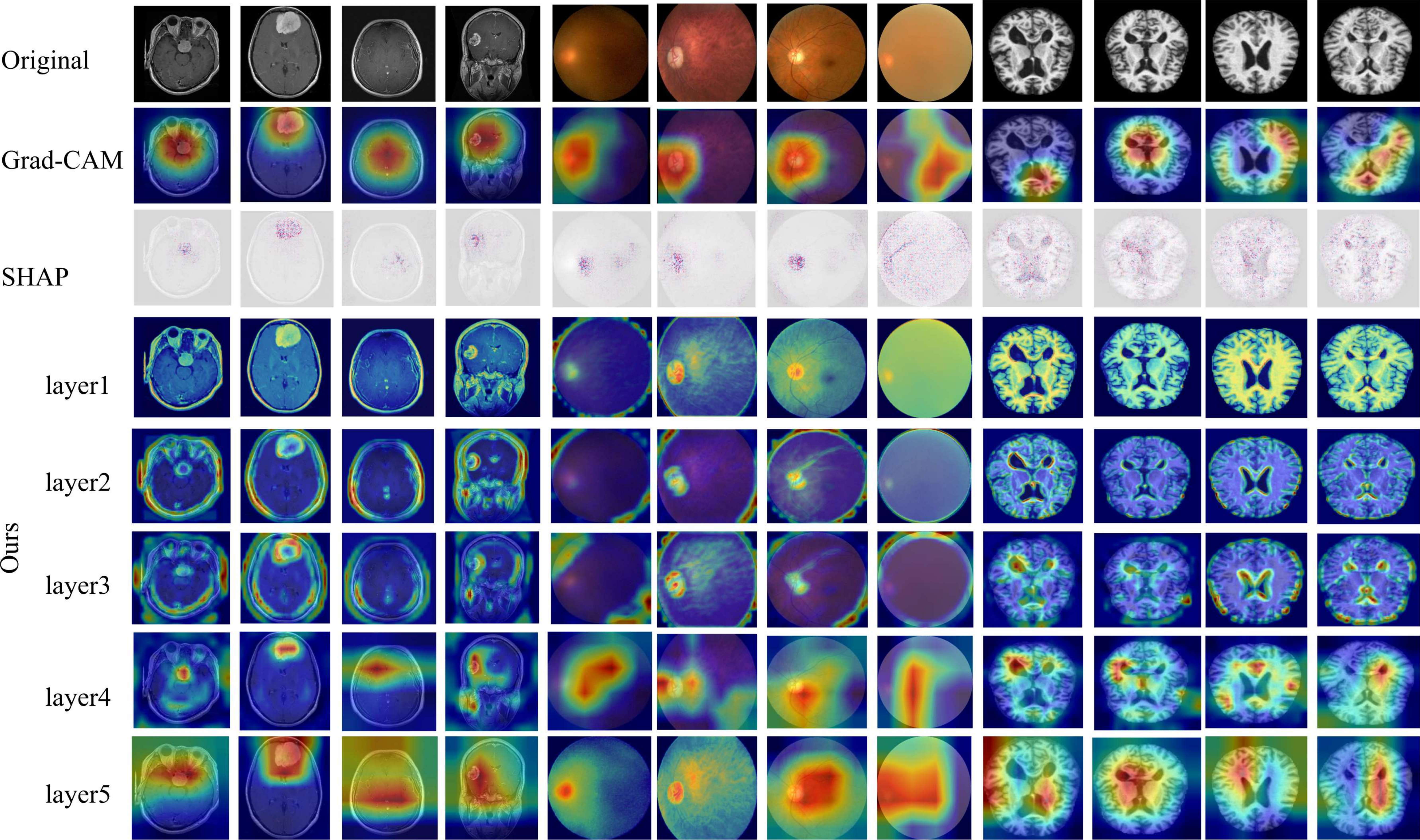}
    \caption{Example of the original, Grad-CAM, SHAP (Teacher) and the proposed method (Student) images that are selected randomly from the test set of each class in the three data sets for interpretability analysis. Specifically, the analyses using Grad-CAM and SHAP rely on teacher models, whereas the proposed method uses student models. The student model has a total of five convolutional layers, and uses the average feature map to view the features of each layer. Each row from left to right represents 12 disease classes from three datasets, namely: Pituitary, Meningioma, No tumor, Glioma, Cataract, Diabetic retinopathy, Glaucoma, Normal, Mild demented, Moderate demented, Non demented, and Very mild demented.}
    \label{XAI}
\end{figure*}

\begin{table}[!ht] \scriptsize
    \normalsize % 设置为10号罗马字体
    \setlength{\tabcolsep}{6.7pt}
    \renewcommand{\arraystretch}{1}
    \dorowcolors
    \resizebox{\columnwidth}{!}{
    \begin{tabular}{cccccc} 
     \toprule
         Model & Accuracy & Loss & $\alpha$ & $T$ & Avg. F1 score \\
           \midrule
         \multicolumn{6}{c}{\textit{Brain tumor}} \\
          \midrule
             DenseNet121 ($T_{B}$) & 0.9877 & 0.0621 & - & - & 0.99\\
             $S_{B1}$ & 0.9572 & 0.1288 & 0.4  & 5 &0.96\\
             $S_{B2}$ & 0.9656 & 0.1297 & 0.4  & 10 &0.97\\
             $S_{B3}$ & 0.9748 & 0.0977 & 0.4  & 15 &0.98\\
             $S_{B4}$ & 0.9702 & 0.1313 & 0.7  & 5 &0.98\\
             $S_{B5}$ & \textbf{0.9748} & \textbf{0.0944} & 0.7  & 10 &0.98\\
             $S_{B6}$ & 0.9717 & 0.1440 & 0.7  & 15 &0.97\\
             \midrule
             \multicolumn{6}{c}{\textit{Eyes disease}} \\
             \midrule
             DenseNet121 ($T_{E}$) & 0.9837 & 0.0490 & - & - & 0.99\\
             $S_{E1}$ & 0.9074 & 0.2468 & 0.4  & 5 & 0.91\\
             $S_{E2}$ & 0.8750 & 0.3510 & 0.4  & 10 & 0.88\\
             $S_{E3}$ & \textbf{0.9351} & \textbf{0.1956} & 0.4  & 15 &0.94\\
             $S_{E4}$ & 0.8958 & 0.3918 & 0.7  & 5 &0.90\\
             $S_{E5}$ & 0.8680 & 0.4633 & 0.7  & 10 &0.87\\
             $S_{E6}$ & 0.9219 & 0.2726 & 0.7  & 15 &0.92\\
             \midrule
             \multicolumn{6}{c}{\textit{Alzheimer}} \\
             \midrule
             DenseNet121 ($T_{A}$) & 0.9938 & 0.0247 & - & - & 0.99\\
             $S_{A1}$ & \textbf{0.9946}& \textbf{0.0194} & 0.4  & 5 & 0.99\\
             $S_{A2}$ & 0.9906 & 0.0274 & 0.4  & 10 & 0.99\\
             $S_{A3}$ & 0.9928 & 0.0220 & 0.4  & 15 &0.99\\
             $S_{A4}$ & 0.9934 & 0.0238 & 0.7  & 5 &0.99\\
             $S_{A5}$ & 0.9915 & 0.0249 & 0.7  & 10 &0.99\\
             $S_{A6}$ & 0.9919 & 0.0288 & 0.7  & 15 &0.99\\
         \bottomrule
    \end{tabular}    }
    \caption{Performance metrics of classifications using three medical datasets. The test accuracy is given by varies $\alpha$ and $T$. The Avg. F1 score is measured by averaging the predicted correct samples for each class. The best student model result is indicated with \textbf{bold} text. $T$ and $S$ denote the $Teacher$ and $Student$ models, respectively. }
   \label{T:1}\vspace{-0.3cm}
\end{table}

\subsection*{Implementation details}
We used an Intel i9-13900k, an NVIDIA GeForce RTX 4090, and TensorFlow-gpu 2.6.0 for our simulations. We used consistent hyperparameter settings, employed the Adam optimizer, set the learning rate to $1 \times 10^{-4}$, and configured the batch size to 16. The model we used as a teacher model is DenseNet121 pretrained on ImageNet \cite{huang2017densely}. In the training phase of the teacher model, average pooling is used to extract features, with the aim of preserving background information and considering overall characteristics effectively. However, for the student model, we used max pooling to capture the most prominent features within an area. The training of the models is performed on a single GPU, while the interpretability tasks are carried out on the CPU. We used standard classification performance metrics such as accuracy, and F-score.

Figure \ref{Fig:CM} shows the confusion matrix obtained from both the teacher model and the student model, which is used for class-wise performance evaluations. We also used the Receiver Operating Characteristic (ROC) curve \cite{mandrekar2010receiver} and Decision Curve Analysis (DCA) \cite{vickers2006decision} to evaluate the classifier models.

\subsection*{Simulation result}
During the training process of the teacher model, we consistently use Densenet121 as our teacher model across each dataset. After obtaining the teacher model, we experiment with different parameters ($\alpha$ and $T$) to perform KD on it, aiming to achieve optimal performance of the student model. The test accuracy, F1 score and test loss of each model are reported in Table \ref{T:1}.

\noindent\textit{Brain Tumor}: The test accuracy of the teacher model can reach 0.9676. Furthermore, the student model is able to perform well in the test set. The best model is the $S_5$ model ($\alpha$=0.7, $T$=10) with the best test accuracy of 0.9748 and test loss of 0.0944.

\noindent\textit{Alzheimer}: As reported in Table \ref{T:1}, the test accuracy of the student model exceeds even that of the teacher model (e.g., 0.9946 vs. 0.9938 using $S_{A1}$ and $T_A$, respectively). This rare situation may be due to the selection of good temperature parameters, which causes the student model to learn useful information faster. In addition, both $T_A$ and all student models can achieve a similar Avg. F1 score ($\sim 0.99$), indicating that the student model can learn rich features from the teacher.

% Based on the validation accuracy, the teacher model consistently maintains a higher accuracy, remaining close to 0.8 throughout the training process. However, a discernible disparity in performance is evident between the cohort of student models and the teacher model, with the former not attaining a level of validation accuracy commensurate with that of the teacher model. %This phenomenon is even more obvious in the validation loss. 
% There is even a gap of nearly 0.5 between the teacher model and the student model.

\noindent\textit{Eyes disease}: As illustrated in Table \ref{T:1}, the best student model is $S_3$ ($\alpha$ = 0.4, $T$ = 15) with a test accuracy of 0.9351 and a test loss of 0.1956. Unlike the BT and Alzheimer datasets, the student model demonstrates lower accuracy compared to $T_E$ (e.g., approximately 5\% decrease). This suggests that the use of shallow networks is limited. Overall, synthesizing performance across the three datasets reveals that KD has enabled the student models to approach the teacher model in terms of validation accuracy. The student model achieved an average F1 score of 0.99, 0.94, and 0.98 in the Alzheimer's disease, Eye disease, and Brain tumor datasets, respectively. Those results are consistent with the confusion matrices of the student and teacher models in all datasets as illustrated in Figure \ref{Fig:CM}.

% Figure \ref{Fig:AUC and ROC} shows the ROC curves of the student and teacher models in the test sets. We observe that the AUC of the student model after distillation is nearly identical to that of the teacher model, and its DCA curve shows almost equivalent training effectiveness.
% \begin{figure}[!ht]
%      \centering
%     \includegraphics[width=0.47\textwidth]{img/SR ROC and DCA.pdf}
%     \caption{DCA (\textbf{first row}) and ROC (\textbf{second row}) curves of Student and Teacher models using three datasets. The left, middle, and right columns represent the brain tumor, eyes disease and Alzheimer datasets, respectively.}
%     \label{Fig:AUC and ROC}
% \end{figure} 

%%%%%%%%%%%%%%%%%%%%%%%%%%%%%%%%%%
%%%%%%%%%%%%%%%%%%%%%%%%%%%%%%%%%%%%%%%%%%

\subsection*{Interpretability analysis}
An interpretability study was conducted to understand the decision-making mechanisms of the classifier model. We extracted an image from each class in the three datasets, selected the best performing student model for layer-by-layer interpretability analysis, and used the teacher model to perform Grad-CAM and SHAP analysis to verify the effectiveness of the proposed method, as shown in Figure \ref{XAI}. Based on the results in Figure \ref{XAI}, clinicians would focus more on the output of the last layer.

\noindent\textit{Grad-CAM}: For \textit{Brain tumor}, 
in the heatmap for Glioma, Meningioma, and Pituitary tumor classes, red highlights areas predicted as tumors on MRI scans, while dark blue marks the background regions of the brain. For the "No tumor" class, there are no designated tumor regions, thus in these heatmaps, the brightest zones are not focused on any specific tumor location due to the absence of an actual tumor. Instead, these highlighted areas may signal different features or regions that the model uses to class the image as "No tumor".
For the \textit{Eyes disease}, except for the "Normal" class, the focus of the heat map is near the optic disc, which is one of the key parts for diagnosis and is related to many eye diseases.
For the \textit{Alzheimer's disease}, in the "Non demented" class, the size of the sulci and ventricles are within the normal range, there is no enlargement and there is no brain atrophy. In “Very mild demented”, the MRI image shows mild ventricular enlargement, the hippocampus shows a shrinking trend, and the focus of the heat map is around the ventricles. In "Mild demented", the sulci may become more pronounced, and the ventricles relatively larger. For the "Moderate demented" class, MRI images showed significant brain atrophy that affected a wider range of brain regions.

\noindent\textit{SHAP}: For understanding, we also performed the SHAP interpretability plots specifically for classes that were predicted accurately. In \textit{Brain tumor} image, there are many SHAP values with positive weights near the tumors. In \textit{Eyes disease} images, SHAP positive weights are distributed around the optic disc, which is similar to the result of heatmap as well. For the normal class, the positive weights are evenly distributed throughout the retina. In \textit{Alzheimer} images, for each class, each feature contributes differently to the prediction of the model, resulting in a wide distribution of SHAP values. In addition, Alzheimer's disease symptoms are associated with many parts of the brain, which can lead the model to identify a variety of influencing factors, resulting in a wide range of SHAP values \cite{bhattarai2024explainable}. In this case, in addition to being able to accurately determine the class of the image, no effective interpretability analysis results can be obtained.

\begin{figure}[t]
     \centering
    \includegraphics[width=0.475\textwidth]{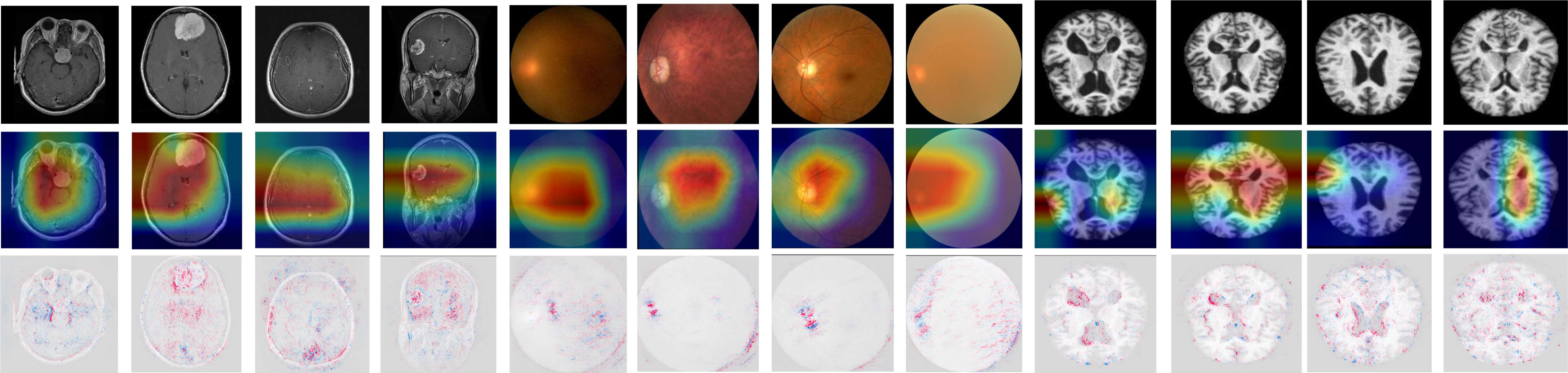}
    \caption{Example of the best student model to perform Grad-CAM and SHAP methods. The first, second and third rows represent the original image, Grad-CAM heatmap and SHAP, respectively. Each column from left to right represents 12 disease classes, namely: Pituitary, Meningioma, No tumor, Glioma, Cataract, Diabetic retinopathy, Glaucoma, Normal, Mild demented, Moderate demented, Non demented, and Very mild demented.}
    \label{XAI2}\vspace{-0.3cm}
\end{figure}

\noindent\textit{CNN features based analysis}: Our student model has only five convolutional layers, which can be approximately regarded as the model inference process with high convolutional layers. For \textit{Brain tumor}, the first few convolutional layers usually focus on the basic elements of the image, such as the basic structure and contours of the brain. As the network level deepens, the middle layers begin to extract more complex features, and from the third layer onward, it gradually changes to learning shapes, specific texture patterns, or components of local objects. Deeper in the network, like the fourth layer, the focus starts to be on regions of the image that are directly related to the predicted class. In the fifth layer, it covers most of the area, including the tumor, which is caused by the maximum pooling layer. Our initial purpose is to reduce the spatial dimension of the feature map while retaining the most important features. This dimensionality reduction operation will lead to the spatial resolution of the feature map. The rate is reduced, so in deep convolutional networks, even small local features may occupy a larger area on the visualized heat map \cite{wang2021avnc}. For \textit{Eyes disease}, at the initial layers, the model primarily detects the contours of the optic disc and the eyes. As the network goes deeper, it increasingly focuses on specific regions of the image, indicating the identification of distinct structures or features within the eyes. In the more advanced layers, the focus of the heat map becomes sharper and more defined. These concentrated areas, or hot spots, typically align with the macula, optic disc, and surrounding blood vessels, which are important for diagnosing retinal conditions. In the "normal" class, the heat map covers nearly the entire image, which indicates an absence of significant anomalies. For \textit{Alzheimer}, like with other datasets, the model initially concentrates on the contours of the brain. In the fourth layer, it becomes apparent that multiple areas are influencing the decision-making process of the model. In the fifth layer, only the segments with the highest impact of the features are retained in each region, aligning with the findings we derived from using Grad-CAM.

\begin{table}[!ht]\small
    \setlength{\tabcolsep}{0.9pt}
    \renewcommand{\arraystretch}{1}
    \resizebox{\columnwidth}{!}{
    \begin{tabular}{ccccc} 
     \toprule
          Brain tumor & Glioma & Meningioma & Pituitary  & Avg \\
          \midrule
   \rowcolor{gray!15}  Grad-CAM & 0.9639 & 0.8934 & \textbf{0.9322} & 0.9298\\ 
    SHAP & 0.9524 & \textbf{0.9687} & 0.9002 & 0.9404\\
   \rowcolor{gray!15}  Ours & \textbf{0.9764} & 0.9161 & 0.9161 & 0.9277 \\
          \midrule
 Eyes disease & Cataract & Diabetic\_retinopathy & Glaucoma & Avg \\
          \midrule
     \rowcolor{gray!15}Grad-CAM & \textbf{0.8555} & \textbf{0.9684} & 0.8939 & 0.9059\\
      SHAP & 0.7733 & 0.9556 & 0.8814 & 0.8700 \\
   \rowcolor{gray!15} Ours & 0.7852 & 0.9080 & \textbf{0.8958} & 0.8630 \\
           \midrule
       \rowcolor{white} Alzheimer & Mild demented & Moderate demented & Very mild demented & Avg \\
          \midrule
      \rowcolor{gray!15}Grad-CAM & \textbf{0.9905} & \textbf{0.9898} & 0.7533 & 0.9112\\
   SHAP & 0.9815 & 0.9666 & \textbf{0.8595} & 0.9359\\
      \rowcolor{gray!15}Ours & 0.9613 & 0.9508 & 0.8274 & 0.9132 \\
         \bottomrule
     \end{tabular}
     }
     \caption{Fidelity score for each class of dataset. The bold values indicate the best results.}
    \label{T:3}\vspace{-0.3cm}
\end{table}

\subsection*{XAI evaluation}
We used the Fidelity score to evaluate the consistency of the explanations generated by the explanation method with the behavior of the original model. A higher Fidelity Score means that the confidence levels of the adversarial and original samples are more similar, indicating that the adversarial sample does not significantly change the model predictions. The given value $C_{a d v}\left(y_{\text {true }}\right)$ represents the confidence of the model (maximum value in the predicted probabilities) in the adversarial sample for the original class $y_{\text {true }}$. $C_{\text {orig }}\left(y_{\text {true }}\right)$ represents the confidence of the model in the original sample for the original class $y_{\text {true }}$. The Fidelity Score is calculated using the following formula:

\begin{equation}\footnotesize
\text { Fidelity score }=\frac{C_{\text {adv}}\left(y_{\text {true }}\right)}{C_{\text {orig }}\left(y_{\text {true }}\right)}
\label{XAI equation}
\end{equation}

Because adversarial generation has minimal impact on the normal class, we focus exclusively on the disease classes. Table \ref{T:3} reports the Fidelity score for the XAI methods. SHAP provides a higher Fidelity score overall, indicating that its interpretative performance is relatively stable and reliable in different cases. Grad-CAM performs well in certain specific classes, but scores lower in others, such as Very Mild Demented. Grad-CAM has relative limitations in detecting minor lesions compared to SHAP, which evaluates the contributions of all features. Although the proposed method achieves feasible performance across classes, on the whole it mimics the impacts of the Grad-CAM and SHAP techniques.

\subsection*{Time-efficient model interpretability }
An additional distinct advantage of the proposed method is that it could reduce the computational effort and the time required. Due to its small number of layers, a quick interpretability analysis can be done on a large number of images. We used two metrics to demonstrate the portability of our model. The first is Floating Point Operations Per Second (FLOPs), which is used to indicate the performance of a computer. It is related to floating-point operations on a computer that can be performed in one second. Another metric is the Mean Execution Time (MET), which indicates the average time it takes for an algorithm to run once. These measurements are reported in Table \ref{T:2}.  Specifically, Table \ref{T:2} shows that the FLOPs value depends solely on the model, with the student model having less than 50\% of the teacher model FLOPs. 

\begin{table}[t]
    \setlength{\tabcolsep}{7.2pt}
    \renewcommand{\arraystretch}{1}
    \fontsize{8.5}{10}\selectfont
    \begin{tabular}{ccccc} 
     \toprule
         \multirow{2}{*}{}&\multicolumn{2}{c}{Grad-CAM} & \multicolumn{2}{c}{SHAP} \\
                 \cline{2-3} \cline{4-5}
        Model & FLOPs$\times10^6$& MET(s) & FLOPs$\times10^6$ & MET(s) \\
         \midrule
       \rowcolor{gray!15} $T$  (BT)& 566.89 & 0.9521 & 566.89 & 67.6837 \\
        $S$  (BT) & 232.70 & 0.1963 & 232.70 & 15.6363 \\
       \rowcolor{gray!15} $T$  (ED)& 566.89 & 0.9389 & 566.89 & 67.5648 \\
        $S$  (ED) & 232.70 & 0.1899 & 232.70 & 15.3973 \\
       \rowcolor{gray!15} $T$  (AD)& 566.89 & 0.9346 &  566.89& 68.4630 \\
        $S$  (AD) & 232.70 & 0.1941 & 232.70 & 15.6316 \\
            \bottomrule
    \end{tabular}
    \caption{Floating point of operations (FLOPs) and mean execution time (MET) of the teacher and student models simulated using brain tumor (BT), eyes disease (ED) and Alzheimer (AD) datasets. $T$: teacher model, $S$: student model, (s): second.}  
    \label{T:2}
\end{table}
This is important for deploying models in resource-limited environments, like mobile devices, as it requires less power and has faster inference times. Various interpretability methods applied show a significant decrease in MET. Grad-CAM provides immediate feedback, producing results in less than one second, requiring only one-fifth of the time of traditional methods while effectively highlighting important lesion regions. The results of the Grad-CAM and SHAP analysis for the student model are shown in Figure \ref{XAI2}.

The MET disparity is more pronounced during SHAP analysis. In medical data sets with many features and interactions, SHAP value calculations can be lengthy. Typically, analyzing a single image with SHAP takes about one minute, but the proposed method reduces this to fifteen seconds, important for efficient large-scale image analysis. SHAP ability to easily identify classes aids in quickly diagnosing patients conditions, allowing faster clinical decisions based on interpretable data. Large hospital-managed data sets with hundreds of classes and millions of images require advanced CNN architectures for feature extraction and classification. However, the proposed method enables the student model to learn from the soft labels of the teacher model, capturing nuanced differences between classes. This helps the student model maintain high accuracy despite reduced complexity.

\section*{Conclusion}\label{sec:5}
In this paper, we presented the intersection of KD and XAI within the CNN model for classification tasks. This approach to simplifying CNN architectures through KD effectively retains essential feature representations while reducing the complexity and size of the model. Experimental results demonstrated that this method maintains high classification accuracy. Furthermore, the use of average feature maps to visualize the focus of the features within each CNN layer allows clinicians to understand the decision-making process of the model. In addition, it reduces computation time and saves computing resources, which is beneficial in high-volume medical image processing. So far, this study presents a faster and more efficient way for healthcare providers to provide better care to patients.

\section*{Acknowledgments}
This research was funded by the National Natural Science Foundation of China grant number 82260360, the Guilin Innovation Platform and Talent Program 20222C264164, and the Guangxi Science and Technology Base and Talent Project (2022AC18004, 2022AC21040).
% \bigskip
% \noindent Thank you for reading these instructions carefully. We look forward to receiving your electronic files!

\bibliography{aaai25}

\begin{thebibliography}{35}
\providecommand{\natexlab}[1]{#1}

\bibitem[{Abhishek et~al.(2023)Abhishek, Jha, Sinha, and Jha}]{abhishek2023automated}
Abhishek, A.; Jha, R.~K.; Sinha, R.; and Jha, K. 2023.
\newblock Automated detection and classification of leukemia on a subject-independent test dataset using deep transfer learning supported by Grad-CAM visualization.
\newblock \emph{Biomedical Signal Processing and Control}, 83: 104722.

\bibitem[{Al-lQubaydhi et~al.(2024)Al-lQubaydhi, Alenezi, Alanazi, Senyor, Alanezi, Alotaibi, Alotaibi, Razaque, and Hariri}]{al2024deep}
Al-lQubaydhi, N.; Alenezi, A.; Alanazi, T.; Senyor, A.; Alanezi, N.; Alotaibi, B.; Alotaibi, M.; Razaque, A.; and Hariri, S. 2024.
\newblock Deep learning for unmanned aerial vehicles detection: A review.
\newblock \emph{Computer Science Review}, 51: 100614.

\bibitem[{Alabi et~al.(2023)Alabi, Elmusrati, Leivo, Almangush, and M{\"a}kitie}]{alabi2023machine}
Alabi, R.~O.; Elmusrati, M.; Leivo, I.; Almangush, A.; and M{\"a}kitie, A.~A. 2023.
\newblock Machine learning explainability in nasopharyngeal cancer survival using LIME and SHAP.
\newblock \emph{Scientific Reports}, 13(1): 8984.

\bibitem[{Amiri et~al.(2024)Amiri, Heidari, Navimipour, Esmaeilpour, and Yazdani}]{amiri2024deep}
Amiri, Z.; Heidari, A.; Navimipour, N.~J.; Esmaeilpour, M.; and Yazdani, Y. 2024.
\newblock The deep learning applications in IoT-based bio-and medical informatics: a systematic literature review.
\newblock \emph{Neural Computing and Applications}, 36(11): 5757--5797.

\bibitem[{Bhattarai et~al.(2024)Bhattarai, Thakuri, Nie, and Chand}]{bhattarai2024explainable}
Bhattarai, P.; Thakuri, D.~S.; Nie, Y.; and Chand, G.~B. 2024.
\newblock Explainable AI-based Deep-SHAP for mapping the multivariate relationships between regional neuroimaging biomarkers and cognition.
\newblock \emph{European Journal of Radiology}, 174: 111403.

\bibitem[{Bhuvaji(2020)}]{BrainTumorClassification}
Bhuvaji, S. 2020.
\newblock Brain Tumor Classification Using Deep Learning Algorithms.
\newblock \url{https://github.com/SartajBhuvaji/Brain-Tumor-Classification-Using-Deep-Learning-Algorithms}.
\newblock Accessed: 2024-12-10.

\bibitem[{Chaddad et~al.(2024)Chaddad, Hu, Wu, Wen, and Kateb}]{chaddad2024generalizable}
Chaddad, A.; Hu, Y.; Wu, Y.; Wen, B.; and Kateb, R. 2024.
\newblock Generalizable and Explainable Deep Learning for Medical Image Computing: An Overview.
\newblock \emph{Current Opinion in Biomedical Engineering}, 100567.

\bibitem[{Chaddad et~al.(2023{\natexlab{a}})Chaddad, Lu, Li, Katib, Kateb, Tanougast, Bouridane, and Abdulkadir}]{chaddad2023explainable}
Chaddad, A.; Lu, Q.; Li, J.; Katib, Y.; Kateb, R.; Tanougast, C.; Bouridane, A.; and Abdulkadir, A. 2023{\natexlab{a}}.
\newblock Explainable, domain-adaptive, and federated artificial intelligence in medicine.
\newblock \emph{IEEE/CAA Journal of Automatica Sinica}, 10(4): 859--876.

\bibitem[{Chaddad et~al.(2023{\natexlab{b}})Chaddad, Peng, Xu, and Bouridane}]{chaddad2023survey}
Chaddad, A.; Peng, J.; Xu, J.; and Bouridane, A. 2023{\natexlab{b}}.
\newblock Survey of explainable AI techniques in healthcare.
\newblock \emph{Sensors}, 23(2): 634.

\bibitem[{Chaddad, Wu, and Desrosiers(2024)}]{chaddad2023federated}
Chaddad, A.; Wu, Y.; and Desrosiers, C. 2024.
\newblock Federated learning for healthcare applications.
\newblock \emph{IEEE Internet of Things Journal}, 11(5): 7339--7358.

\bibitem[{Doddi(2023)}]{doddi2023eye}
Doddi, G.~V. 2023.
\newblock Eye Diseases Classification.
\newblock https://www.kaggle.com/datasets/gunavenkatdoddi/eye-diseases-classification.

\bibitem[{Figshare(2017-04-03)}]{BrainTumorDataset}
Figshare. 2017-04-03.
\newblock Brain Tumor Dataset.
\newblock \url{https://figshare.com/articles/dataset/brain_tumor_dataset/1512427}.
\newblock Accessed: 2024-12-10.

\bibitem[{Gou et~al.(2021)Gou, Yu, Maybank, and Tao}]{gou2021knowledge}
Gou, J.; Yu, B.; Maybank, S.~J.; and Tao, D. 2021.
\newblock Knowledge distillation: A survey.
\newblock \emph{International Journal of Computer Vision}, 129(6): 1789--1819.

\bibitem[{Hamada(2020)}]{br35h2020}
Hamada, A. 2020.
\newblock BR35H Brain Tumor Detection 2020.
\newblock https://paperswithcode.com/dataset/br35h-brain-tumor-detection-2020.

\bibitem[{Hassija et~al.(2024)Hassija, Chamola, Mahapatra, Singal, Goel, Huang, Scardapane, Spinelli, Mahmud, and Hussain}]{hassija2024interpreting}
Hassija, V.; Chamola, V.; Mahapatra, A.; Singal, A.; Goel, D.; Huang, K.; Scardapane, S.; Spinelli, I.; Mahmud, M.; and Hussain, A. 2024.
\newblock Interpreting black-box models: a review on explainable artificial intelligence.
\newblock \emph{Cognitive Computation}, 16(1): 45--74.

\bibitem[{Hinton, Vinyals, and Dean(2015)}]{hinton2015distilling}
Hinton, G.; Vinyals, O.; and Dean, J. 2015.
\newblock Distilling the knowledge in a neural network.
\newblock \emph{arXiv preprint arXiv:1503.02531}.

\bibitem[{Huang et~al.(2017)Huang, Liu, Van Der~Maaten, and Weinberger}]{huang2017densely}
Huang, G.; Liu, Z.; Van Der~Maaten, L.; and Weinberger, K.~Q. 2017.
\newblock Densely connected convolutional networks.
\newblock In \emph{Proceedings of the IEEE conference on computer vision and pattern recognition}, 4700--4708.

\bibitem[{Jin et~al.(2023)Jin, Li, Fatehi, and Hamarneh}]{jin2023guidelines}
Jin, W.; Li, X.; Fatehi, M.; and Hamarneh, G. 2023.
\newblock Guidelines and evaluation of clinical explainable AI in medical image analysis.
\newblock \emph{Medical Image Analysis}, 84: 102684.

\bibitem[{Kaggle(2022)}]{AugmentedAlzheimerMRIDataset}
Kaggle. 2022.
\newblock Augmented Alzheimer MRI Dataset.
\newblock \url{https://www.kaggle.com/datasets/uraninjo/augmented-alzheimer-mri-dataset}.
\newblock Accessed: 2024-12-10.

\bibitem[{Khan et~al.(2022)Khan, Alam, Dhruba, Zunair, and Mohammed}]{khan2022knowledge}
Khan, M.~S.; Alam, K.~N.; Dhruba, A.~R.; Zunair, H.; and Mohammed, N. 2022.
\newblock Knowledge distillation approach towards melanoma detection.
\newblock \emph{Computers in Biology and Medicine}, 146: 105581.

\bibitem[{Liu et~al.(2023)Liu, Desrosiers, Ayed, and Dolz}]{liu2023segmentation}
Liu, B.; Desrosiers, C.; Ayed, I.~B.; and Dolz, J. 2023.
\newblock Segmentation with mixed supervision: Confidence maximization helps knowledge distillation.
\newblock \emph{Medical Image Analysis}, 83: 102670.

\bibitem[{Lundberg and Lee(2017)}]{lundberg2017unified}
Lundberg, S.~M.; and Lee, S.-I. 2017.
\newblock A unified approach to interpreting model predictions.
\newblock \emph{Advances in neural information processing systems}, 30.

\bibitem[{Mandrekar(2010)}]{mandrekar2010receiver}
Mandrekar, J.~N. 2010.
\newblock Receiver operating characteristic curve in diagnostic test assessment.
\newblock \emph{Journal of Thoracic Oncology}, 5(9): 1315--1316.

\bibitem[{Nickparvar(2021)}]{nickparvar2021brain}
Nickparvar, M. 2021.
\newblock Brain Tumor MRI Dataset.
\newblock \url{https://www.kaggle.com/datasets/masoudnickparvar/brain-tumor-mri-dataset}.
\newblock Accessed: 2024-12-11.

\bibitem[{Park et~al.(2022)Park, Kim, Oh, Seo, Lee, Kim, Moon, Lim, Park, and Ye}]{park2022self}
Park, S.; Kim, G.; Oh, Y.; Seo, J.~B.; Lee, S.~M.; Kim, J.~H.; Moon, S.; Lim, J.-K.; Park, C.~M.; and Ye, J.~C. 2022.
\newblock Self-evolving vision transformer for chest X-ray diagnosis through knowledge distillation.
\newblock \emph{Nature communications}, 13(1): 3848.

\bibitem[{Patel et~al.(2023)Patel, Aggarwal, Solanki, Dahiya, Yadav et~al.}]{patel2023logistic}
Patel, R.~K.; Aggarwal, E.; Solanki, K.; Dahiya, O.; Yadav, S.~A.; et~al. 2023.
\newblock A Logistic Regression and Decision Tree Based Hybrid Approach to Predict Alzheimer's Disease.
\newblock In \emph{2023 International Conference on Computational Intelligence and Sustainable Engineering Solutions (CISES)}, 722--726. IEEE.

\bibitem[{Raihan et~al.(2023)Raihan, Khan, Kee, and Nahid}]{raihan2023detection}
Raihan, M.~J.; Khan, M. A.-M.; Kee, S.-H.; and Nahid, A.-A. 2023.
\newblock Detection of the chronic kidney disease using XGBoost classifier and explaining the influence of the attributes on the model using SHAP.
\newblock \emph{Scientific Reports}, 13(1): 6263.

\bibitem[{Selvaraju et~al.(2017)Selvaraju, Cogswell, Das, Vedantam, Parikh, and Batra}]{selvaraju2017grad}
Selvaraju, R.~R.; Cogswell, M.; Das, A.; Vedantam, R.; Parikh, D.; and Batra, D. 2017.
\newblock Grad-cam: Visual explanations from deep networks via gradient-based localization.
\newblock In \emph{Proceedings of the IEEE international conference on computer vision}, 618--626.

\bibitem[{Song et~al.(2023)Song, Yao, Jiang, Shi, Cui, Wang, Wang, Wu, Tian, Ye et~al.}]{song2023new}
Song, D.; Yao, J.; Jiang, Y.; Shi, S.; Cui, C.; Wang, L.; Wang, L.; Wu, H.; Tian, H.; Ye, X.; et~al. 2023.
\newblock A new xAI framework with feature explainability for tumors decision-making in Ultrasound data: comparing with Grad-CAM.
\newblock \emph{Computer Methods and Programs in Biomedicine}, 235: 107527.

\bibitem[{Termritthikun et~al.(2023)Termritthikun, Umer, Suwanwimolkul, Xia, and Lee}]{termritthikun2023explainable}
Termritthikun, C.; Umer, A.; Suwanwimolkul, S.; Xia, F.; and Lee, I. 2023.
\newblock Explainable knowledge distillation for on-device chest x-ray classification.
\newblock \emph{IEEE/ACM Transactions on Computational Biology and Bioinformatics}.

\bibitem[{Van~der Velden et~al.(2022)Van~der Velden, Kuijf, Gilhuijs, and Viergever}]{van2022explainable}
Van~der Velden, B.~H.; Kuijf, H.~J.; Gilhuijs, K.~G.; and Viergever, M.~A. 2022.
\newblock Explainable artificial intelligence (XAI) in deep learning-based medical image analysis.
\newblock \emph{Medical Image Analysis}, 79: 102470.

\bibitem[{Vickers and Elkin(2006)}]{vickers2006decision}
Vickers, A.~J.; and Elkin, E.~B. 2006.
\newblock Decision curve analysis: a novel method for evaluating prediction models.
\newblock \emph{Medical Decision Making}, 26(6): 565--574.

\bibitem[{Wang et~al.(2023{\natexlab{a}})Wang, Fu, Du, Gao, Huang, Liu, Chandak, Liu, Van~Katwyk, Deac et~al.}]{wang2023scientific}
Wang, H.; Fu, T.; Du, Y.; Gao, W.; Huang, K.; Liu, Z.; Chandak, P.; Liu, S.; Van~Katwyk, P.; Deac, A.; et~al. 2023{\natexlab{a}}.
\newblock Scientific discovery in the age of artificial intelligence.
\newblock \emph{Nature}, 620(7972): 47--60.

\bibitem[{Wang et~al.(2021)Wang, Fernandes, Zhu, and Zhang}]{wang2021avnc}
Wang, S.-H.; Fernandes, S.~L.; Zhu, Z.; and Zhang, Y.-D. 2021.
\newblock AVNC: attention-based VGG-style network for COVID-19 diagnosis by CBAM.
\newblock \emph{IEEE Sensors Journal}, 22(18): 17431--17438.

\bibitem[{Wang et~al.(2023{\natexlab{b}})Wang, Wang, Cai, Lee, Miao, and Wang}]{wang2023ssd}
Wang, Y.; Wang, Y.; Cai, J.; Lee, T.~K.; Miao, C.; and Wang, Z.~J. 2023{\natexlab{b}}.
\newblock Ssd-kd: A self-supervised diverse knowledge distillation method for lightweight skin lesion classification using dermoscopic images.
\newblock \emph{Medical Image Analysis}, 84: 102693.

\end{thebibliography}

\end{document}